%% file: explanation_paper.tex
\documentclass[twoside]{article}

\usepackage[accepted]{aistats2022}
\usepackage{latexsym}

\usepackage[T1]{fontenc}

\usepackage[utf8]{inputenc}
\usepackage{array,multirow}
\usepackage{microtype}

\usepackage{caption}
\usepackage{graphicx}
\usepackage{amsmath}
\usepackage{amsthm}
\usepackage{booktabs}
\usepackage{algorithm}
\usepackage{algorithmic}

\usepackage{subcaption}
\graphicspath{{fig/}}
\usepackage{bbm}
\usepackage{amsfonts}

\newenvironment{flushitemize}{%
\begin{list}{$\bullet$}
   {\setlength{\leftmargin}{15pt}}%
    \setlength{\labelwidth}{20pt}
    \setlength{\itemindent}{7pt}
    \setlength{\labelsep}{0.5em}
 \setlength{\itemsep}{1pt}
 \setlength{\parskip}{0pt}
 \setlength{\parsep}{0pt}}
 {\end{list}}

\usepackage[round]{natbib}

\newtheorem{theorem}{Theorem}

\usepackage{xcolor}

\begin{document}
\twocolumn[

\aistatstitle{Cross-Loss Influence Functions to Explain Deep Network Representations}

\aistatsauthor{ Andrew Silva \And Rohit Chopra \And Matthew Gombolay}
\aistatsaddress{ Georgia Institute of Technology \And Georgia Institute of Technology \And Georgia Institute of Technology}

]

\begin{abstract}

As machine learning is increasingly deployed in the real world, it is paramount that we develop the tools necessary to analyze the decision-making of the models we train and deploy to end-users. Recently, researchers have shown that influence functions, a statistical measure of sample impact, can approximate the effects of training samples on classification accuracy for deep neural networks. However, this prior work only applies to supervised learning, where training and testing share an objective function. No approaches currently exist for estimating the influence of unsupervised training examples for deep learning models. To bring explainability to unsupervised and semi-supervised training regimes, we derive the first theoretical and empirical demonstration that influence functions can be extended to handle mismatched training and testing (i.e., ``cross-loss'') settings. Our formulation enables us to compute the influence in an unsupervised learning setup, explain cluster memberships, and identify and augment biases in language models. Our experiments show that our cross-loss influence estimates even exceed matched-objective influence estimation relative to ground-truth sample impact.
\end{abstract}


\section{INTRODUCTION}
\label{sec:intro}

Deep learning has become a ubiquitous tool in domains from speech-recognition~\citep{bengio2014word} to knowledge-discovery~\citep{tshitoyan2019unsupervised}, owing to large datasets, increasing computation power, and the development of powerful unsupervised and semi-supervised learning techniques. With the proliferation of these models in the real world, it is evermore vital that researchers are able to interrogate their learned models and understand their training data to prevent erroneous or malicious models from being released~\citep{doshi2017towards}. In this work, we derive the first quantitative method to interrogate and augment the decision-making of deep learning models trained without explicitly labeled data, e.g. for self-supervised or unsupervised pre-training. 

To quantitatively explain the impact of training data, \emph{influence functions} have emerged as a powerful tool~\citep{koh2017understanding}. Influence functions provide an estimate of the ``influence'' of every training example on a trained model's performance for a test example.
 Unfortunately, this prior work cannot be applied when a model is trained via one objective and tested with another, such as in unsupervised or self-supervised applications (e.g., clustering or masked-language modeling \citep{vaswani2017attention}). If these systems exhibit sub-optimal decision-making or even bias~\citep{bolukbasi2016man,lum2016predict}, researchers must be able to understand and rectify a model's erroneous decision-making. With such ``cross-loss'' paradigms becoming ubiquitous and seeing real-world deployments, we critically need methods to explain these models and their datasets.

\begin{figure*}
  \centering
  \includegraphics[width=.95\linewidth]{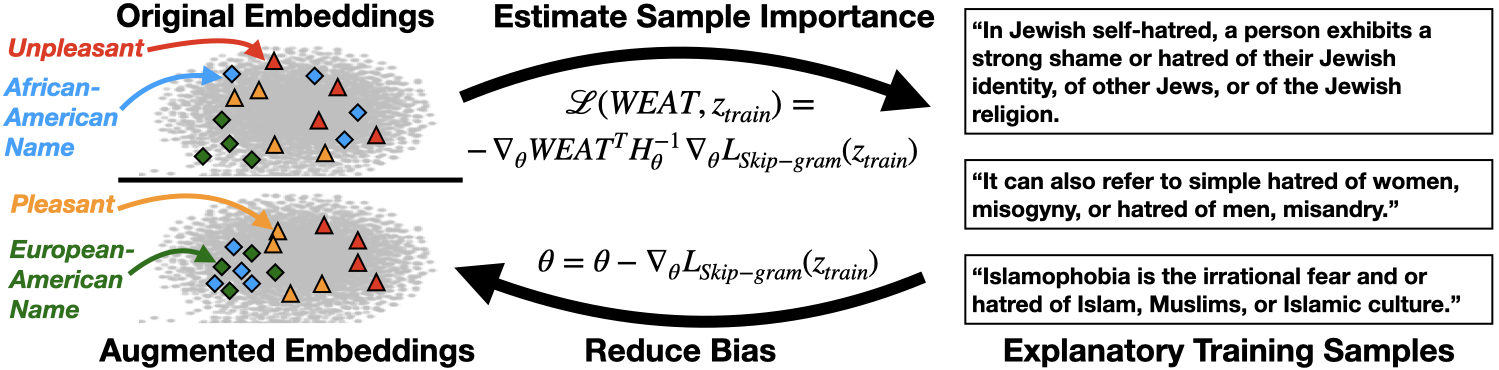}
  \caption{Taking a word embedding model and its training data, our approach explains properties of word embeddings (e.g., a high score on a racial bias test) by identifying which training samples are contributing most to these properties. We employ these training samples to augment the model. The relationship between \textit{Race} and \textit{Religion} is explored in Section \ref{subsubsec:bias-case-study}.}
  \label{fig:pipeline}

\end{figure*}
In this work, we present cross-loss influence functions (CLIF), a novel extension of influence functions that permits use across separate training and testing objectives. Our formulation enables researchers to understand a training sample's effect on deep network representations, even when these models are learned without explicit labeling. We demonstrate the power of our approach on a synthetic dataset and a set of word embedding models, demonstrating that CLIF correlates with ground-truth influence as reliably as matched-objective influence functions. Further, we show that CLIF can be used to identify sequences that contribute to arbitrary properties of learned word embeddings, such as cluster membership or bias scores.  Our novel extension to influence functions contributes a powerful tool for interrogating large datasets, and generalizes the technique to an array of modern deep learning deployments. 

We graphically depict the novel capability of our approach for a candidate application to explainability and debiasing in Figure \ref{fig:pipeline}. Upon finding racial bias in a word embedding model, our approach takes the trained word embedding model with the original dataset and identifies the individual training sequences which contribute most significantly to the racial bias. With the most relevant sequences identified, we can fine-tune the model and undo the influence of these sequences, mitigating the negative bias.

\section{RELATED WORK}
\label{sec:related-work}

\paragraph{Explainability} Explainable machine learning is a growing area of research seeking to help researchers and practitioners alike ensure the desired functions of their machine learning models. As the field receives increased attention, various explainability mechanisms have been proposed, including interpretable neural network approaches~\citep{wang2016bayesian}, natural language generation of explanations~\citep{mullenbach2018explainable,wiegreffe2019attention,chen2021generate}, and many others \cite{hoffman2018metrics,holzinger2020measuring,linardatos2021explainable}.
One promising direction is case-based reasoning or explanation by example~\citep{caruana1999case,gaams}, in which a model returns an explanatory sample for each model prediction. Our CLIF approach affords explanation by example, finding training samples to explain model properties.

In large datasets, finding relevant samples for explanation is not a trivial task, as similarity metrics reveal related data but not explanatory data. \citet{koh2017understanding} introduced the \emph{influence function} to deep networks as a means of discovering the magnitude of each training sample's impact on a deep network's final prediction for an unseen test case. 
The approach of \citet{brunet2018understanding} extends influence functions to GloVe~\citep{pennington2014glove}; however, this method 
is designed specifically for GloVe embeddings with co-occurrence matrix rather than neural networks with arbitrary pre-training objectives. \citet{han-etal-2020-explaining} apply the influence function to supervised sentiment analysis and natural language inference tasks, again being classification tasks with matched train and test objectives. Finally, \citet{pmlr-v108-barshan20a} introduce a normalization scheme for influence functions to improve robustness to outliers. 

Our work is unique in that we are the first to generalize influence functions to mismatched training and testing loss functions. Our approach enables researchers to take a model trained without supervision, such as deep unsupervised learning \citep{xie2016unsupervised} or word embedding models, and explain relationships between embeddings.

\paragraph{Bias in Word Embeddings}  Word embeddings have emerged as a standard for representing language in neural networks, and are therefore ubiquitous in language-based systems. A standard approach to learn these embeddings is Word2Vec~\citep{mikolov2013efficient} which transforms words from one-hot dictionary indices into information-rich vectors. Research has shown that Word2Vec embeddings capture a word's meaning and offer powerful representations of words used in many settings~\citep{bakarov2018survey}. Transformer-based models, such as BERT \citep{devlin2018bert}, build contextual sequence representations by applying attention mechanisms to word-embeddings, leveraging information in the embeddings.


\citet{bolukbasi2016man} were among the first to discover the inherent biases in word embeddings, a finding that has been replicated in 
transformers~\citep{basta2019evaluating,kurita2019measuring,silva2021towards} and more~\citep{ws-2019-gender,swinger2019biases,nissim-etal-2020-fair}. \citet{caliskan2017semantics} presented the Word Embedding Association Test (WEAT) to test word embeddings for implicit bias, e.g. by aligning \textit{male} with \textit{math} and \textit{female} with \textit{arts}.
Our definition of bias and its implications is given in Section \ref{sec:societal-impact}.

\section{APPROACH}
\label{sec:approach}
The aim of our work is to help explain the properties of representations learned via any objectives (e.g., unsupervised or semi-supervised learning) by finding training examples which are most responsible for those properties. Here, we consider an embedding model parameterized by $\theta$. For every test example, $z_{te}$, we want to find the effect of increasing the weight of a training sample $z_{tr}$ on the model's performance on $z_{te}$. \citet{koh2017understanding} show that the effect of upweighting $z_{tr}$ is given by Equation \ref{eqn:influence-function}.
\begin{equation}
    \label{eqn:influence-function}
    \mathcal{L}(z_{te}, z_{tr}) =-\nabla_\theta L(z_{te}, \hat{\theta})^T H_{\hat{\theta}}^{-1} \nabla_\theta L(z_{tr}, \hat{\theta})
\end{equation}
$H_{\hat{\theta}}$ is the Hessian of the loss function with respect to model parameters $\theta$. \citet{koh2017understanding} leverage recent work \citep{agarwal2016second} to efficiently estimate the inverse Hessian-vector product (HVP).

\subsection{Explaining Deep Network Representations}
\label{subsec:approach-explain}
The original influence function from Equation \ref{eqn:influence-function} assumes that the loss function remains the same for both sets of samples $z_{te}$ and $z_{tr}$. We instead seek to compute influence on representations learned by any deep learning models, even if the training and testing tasks are different. To compute the effects of training samples on arbitrary properties of the model's learned representation, e.g.  cluster centroids or scores on a bias test, we need an approach that can compute influence across multiple different objectives. 

To address this need for cross-loss influence estimation, we relax the assumption that both loss functions must be the same. Building on prior work \citep{agarwal2016second}, we estimate the HVP via \textit{any} twice-differentiable objective, thus providing the overall loss landscape for deep network representations as they are refined. This approach is in contrast to prior uses of the influence function for deep networks, which estimated the HVP according to a supervised-learning objective that matched the test task.

We now provide analytical proof for this approach to obtaining explanations for mismatched objectives. First, we review the original influence function formulation, using a training objective
to estimate the effect of each sample on the model parameters. Second, we prove a novel theorem enabling us to substitute any differentiable testing loss function in our cross-loss influence function rather than requiring testing and training loss functions to be equivalent. 

\begin{theorem}
\label{theorem:cross-loss-influence}
Exchanging the testing loss function of Equation \ref{eqn:influence-function} for an arbitrary differentiable objective, we can compute the influence of training examples on the new objective.
\end{theorem}
\begin{proof}
\citet{koh2017understanding} derive that the effect of upweighting a single sample, $z$, on models parameterized by $\hat\theta$ is defined by Equation  \ref{eqn:influence-original}.
\begin{equation}
\label{eqn:influence-original}
\ensuremath{\mathcal{I}}_\text{up,params}(z) := \frac{d\hat\theta_{\epsilon,z}}{d\epsilon}\Bigr|_{\substack{\epsilon = 0}} = -H_{\hat\theta}^{-1} \nabla L(z, \hat\theta)
\end{equation}
Here, $\hat\theta$ is the model parameterization that minimizes the loss $L$. Equation \ref{eqn:influence-original} models the impact of taking an infinitesimally small stochastic gradient descent step evaluated at the test data point, $z$. The updated parameters, $\hat{\theta}_{\epsilon ,z}$, are given by the satisfaction of Equation \ref{eqn:eps-model}.
\begin{align}
\label{eqn:eps-model}
  \hat\theta_{\epsilon, z} := \arg\min_{\theta \in \Theta} \frac{1}{N} \sum_{i=1}^N L(z_i, \theta) + \epsilon L(z, \theta)
\end{align}
For $\hat\theta_{\epsilon, z}$, we can estimate the rate of change for a loss function on a new sample $dL(z_{te}, \hat\theta_{\epsilon, z})$ with respect to the original model parameters $\theta$ by combining the derivative of the loss function with the rate of change for $\hat\theta_{\epsilon, z}$ as a function of $\epsilon$, as in Equation \ref{eqn:new-loss}.
\begin{align}
\label{eqn:new-loss}
\ensuremath{\mathcal{I}}_\text{up}L(z, z_{te}) & := \frac{dL(z_{te}, \hat\theta_{\epsilon, z})} {d\epsilon} \Bigr|_{\substack{\epsilon = 0}} \nonumber \\ 
& = \left[\frac{dL(z_{te}, \hat\theta)} {d\hat\theta} \frac{d\hat\theta_{\epsilon, z}} {d\epsilon} \right]_{\substack{\epsilon = 0}}
\end{align}

It follows, then, that the influence of upweighting a sample $z$ on a different loss function $L'$ is given in Equation \ref{eqn:new-loss-prime}.
\begin{align}
\label{eqn:new-loss-prime}
\ensuremath{\mathcal{I}}_\text{up}L'(z, z_{te}) &:= \frac{dL'(z_{te}, \hat\theta_{\epsilon, z})} {d\epsilon} \Bigr| _{\substack{ \epsilon=0 }} \nonumber \\ 
&= \left[ \frac{dL'(z_{te}, \hat\theta)} {d\hat\theta} \frac{d\hat\theta_{\epsilon, z}} {d\epsilon} \right]_{\substack{\epsilon = 0}}
\end{align}
This derivation indicates that we can estimate the influence of changing parameters on a new loss function $L'$, as the influence is determined as a function of $\theta$, which exists as the current model parameters. 
To estimate influence, we want to estimate how the value of $L'$ will change as $\hat\theta$ is modified by $\epsilon$-sized gradient steps (i.e., how would upweighting a sample affect the final $L'$?). The rate of change of this effect is given as $\frac{d\hat\theta_{\epsilon, z}}{d\epsilon}$. Finally, we can estimate this change given Equation \ref{eqn:influence-original} with our training loss $L$.
\end{proof}
This result proves that influence functions can be extended to \textit{any} problem with a twice-differentiable training objective, opening up a wide array of new domains for explanation with influential examples. In the remainder of the paper, we empirically validate this analytical result, and present example scenarios for deployment of our CLIF formulation.

\section{EXPERIMENTS}
\label{sec:experiments}

We conduct four experiments utilizing our approach to explain deep network representations. The first is validation of our approach for explainability on a synthetic dataset, verifying our results with leave-one-out retraining and straightforward data visualization. Next, we apply our approach using a small language dataset which we can fully review, allowing us to closely examine select examples as a case-study (Section \ref{subsec:scifi-exp}). Third, we conduct a thorough test of our approach on a dataset which a human could not reasonable review manually, which we additionally follow-up with an in-depth exploration of spurious correlations that our method can discover, but that human annotators would likely miss (Section \ref{subsubsec:bias-case-study}). Finally, we present a novel approach to word-embedding augmentation for bias mitigation using samples discovered by our approach (Section \ref{subsec:bias-mitigation}). 




\subsection{Clustering Accuracy: Synthetic Data}
\label{subsec:synthetic-exp}
In this section, we to demonstrate that CLIF can uncover influential training data for models trained via unsupervised learning (i.e., clustering). In this experiment, we apply Deep Embedded Clustering (DEC) \citep{xie2016unsupervised} to a synthetic dataset to learn a lower-dimensional representation of the data in the form of clusters. We then change contexts to consider a supervised learning loss, $L_{NLL}$, the negative log-likelihood loss for predicting the correct cluster, $y$, given a data point, $z$. This experiment serves as an ideal test domain for CLIF as we are able to empirically validate all influence predictions over the entire dataset in addition to comparing to a matched-objective influence estimation.

 

\begin{figure}[t]
\centering
\begin{subfigure}[b]{\linewidth}
\centering
        \includegraphics[width=\textwidth]{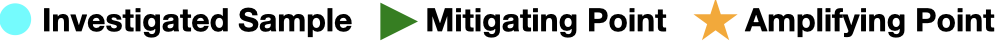}
\end{subfigure}
    \begin{subfigure}[b]{0.45\linewidth}
        \includegraphics[width=\linewidth]{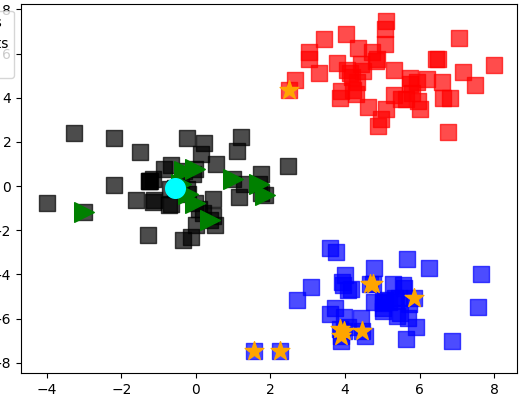}
        \caption{Empirical MOG Influence}
        \label{fig:mog-empirical}
    \end{subfigure}
    ~~
    \begin{subfigure}[b]{0.45\linewidth}
        \includegraphics[width=\linewidth]{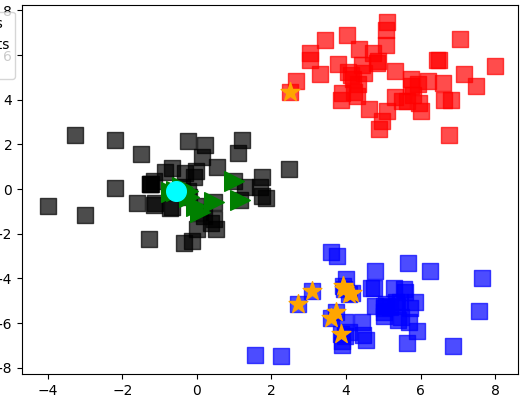}
        \caption{Predicted MOG Influence}
        \label{fig:mog-influence}
    \end{subfigure}
\caption{(Left) After leave-one-out retraining with the entire dataset, we identify the ten most influential samples which amplify and mitigate classification loss in our MOG dataset. (Right) We show the ten most amplifying or mitigating samples according to CLIF.}
\label{fig:mog}
\end{figure}

In our experiment, we train a model to learn three cluster centroids on a mixture of gaussians (MOG) dataset, shown in Figure \ref{fig:mog}. With the synthetic dataset, we train two models-- one via supervised learning (using the $L_{NLL}$ objective), and one via DEC. With these two models, we compare two approaches to influence estimation: (1) using the original influence function, with $L_{NLL}$ as the training and testing loss function (Equation \ref{eqn:nll-influence-function}), and (2) using DEC (ignoring the ground truth labels in training) and testing with $L_{NLL}$(Equation \ref{eqn:dec-influence-function}). The testing loss is evaluated with respect to a prediction, $\hat{y}$, and ground-truth, $y$, for data points, $z$, with training labels, $\hat{z}$.
\begin{equation}
\label{eqn:nll-influence-function}
    \mathcal{L} = -\nabla_\theta L_{NLL}(\hat{y}, y) ^T H_{\theta}^{-1} \nabla_\theta L_{NLL}(\hat{z}, \theta(z))
\end{equation}
\begin{equation}
\label{eqn:dec-influence-function}
    \mathcal{L} = -\nabla_\theta L_{NLL}(\hat{y}, y) ^T H_{\theta}^{-1} \nabla_\theta L_{DEC}(\theta(z))
\end{equation}
This comparison allows us to establish the correlation between the actual, empirical influence of each sample and the predicted influence with CLIF. 

For each data point in this experiment, we ask the question: ``Which points in the dataset have contributed to this sample being assigned to the correct or incorrect class?'' We provide an example of the estimated and actual influence on a single point in Figure \ref{fig:mog}, showing the efficacy of our approach.

 We pass all samples through the influence function in addition to performing leave-one-out (LOO) retraining across the entire dataset, allowing us to compare the predicted vs. actual influence of every point in the dataset on every other point, and obtain a Pearson's correlation coefficient between predicted and actual influence for all samples. 
 
 With matched train and test objectives (i.e. Equation \ref{eqn:nll-influence-function}), we observe one class in which 100\% Pearson correlation coefficients for LOO retraining are $> 0.8$ (i.e. 100\% $>0.8$). We also see strong performance in the remaining classes (38\% $> 0.8$ and 36\% $> 0.8$). When applying CLIF (i.e. Equation \ref{eqn:dec-influence-function}), we see a \textbf{an even stronger} correlation, with one class again achieving near-perfect correlation (100\% $>0.8$), and the remaining classes correlating more closely than matched-objective predictions (52\% $>0.8$ and 44\% $>0.8$).
 \textbf{This result demonstrates the utility of CLIF as the cross-loss influence correlation is higher than the matched-objective influence correlation.}

\subsection{Nearest Neighbors: Science Fiction}
\label{subsec:scifi-exp}

We next evaluate the efficacy of CLIF in explaining word embedding nearest-neighbor relationships for a science fiction dataset, $D_{SCI}$~\citep{ammanabrolu-etal-2019-guided}, bringing our synthetic-dataset experiments to real text data. This dataset includes over 2,000 plot summaries from a handful of science fiction television programs. We expect 
CLIF will be able to find salient sentences in the plot summaries which explain an embedding's local neighborhood.

Throughout our word embedding experiments, we leverage the Skip-gram model, which is based on the idea that words nearby in a sentence are more related than distant words. Each word, $\vec{w}_j$, in a sequence  is paired with its neighbors in a context window to create a set of $k$ context-word pairs $[\vec{c}_1, \vec{w}_j], [\vec{c}_2, \vec{w}_j], ... [\vec{c}_k, \vec{w}_j]$. One also draws a set of unrelated words, $N_i$, from the entire corpus according to their frequency to serve as negative examples.
Each word $\vec{w}_j$, context $\vec{c}$, and negative-set, $N_i$, tuple $\langle \vec{w}_j, \vec{c}, N_i \rangle$ is then an input to the neural network, which is tasked with minimizing the difference between $\vec{w}_j$ and $\vec{c}$ and maximizing the difference between $\vec{w}_j$ and all negative samples in $N_i$, as shown in Equation \ref{eqn:skipgram-loss}, where $\sigma$ is the sigmoid function. 
\begin{equation}
    \label{eqn:skipgram-loss} 
    L_{SG}(\vec{w}_j, \vec{c}, N_i) = \displaystyle \mathop{\mathbb{E}}_{\vec{n} \sim N_i}[\sigma(\vec{w}^T_j \vec{n})] - \sigma(\vec{w}^T_j \vec{c})
\end{equation}

After training a set of Skip-gram word embeddings on $D_{SCI}$, we then examine nearest-neighbors for different words in the dataset, and apply CLIF to discover more about the these nearest-neighbor relationships.
We leverage CLIF with a mean-squared error (MSE) loss modification, given in Equation \ref{eqn:mse-influence-function}. The MSE loss function is the difference between where an embedding $e$ was initialized, $E_{I}(e)$, and where it finishes training, $E_{F}(e)$, which models the question ``Which training samples are most responsible for moving $e$ to its final location/neighborhood?''
\begin{equation}
\label{eqn:mse-influence-function}
    \mathcal{L}(e, z_{t}) =  -\nabla_\theta MSE(e) ^T H_{\theta}^{-1} \nabla_\theta L_{SG}(z_{t})
\end{equation}

\begin{table*}
  \caption{Influential Training Data Explaining Sci-Fi Nearest Neighbors.}

  \begin{tabular}{c|l}
  \hline
  & \textbf{Dooku} \\
    &  ``... Wars, such as seeking Dooku as a place holder Sith apprentice and ally in the war.''\\
    Reinforcing  &  ``the Sith sees will be easily manipulated''\\
    \cline{2-0}
     Excerpts, $A$  &  \textbf{Doctor} \\
     & ``Using the psychic paper, the Doctor presents credentials''\\
    & ``The Doctor begins scanning the house with the sonic screwdriver''\\
    \cline{2-0}
    & \textbf{Kirk} \\
    & ``A landing party from the USS Enterprise comprised of Captain Kirk, Scott, Dr. McCoy, ...'' \\
    & ``Due to this transporter accident, Kirk has been split into two beings'' \\
\specialrule{.2em}{.2em}{.2em}
  
    & \textbf{Dooku} \\
    & ``The young Padawan Dooku had recently been appointed to Jedi Thame Cerulian.''\\ 
  Removing  & ..now a full-pledged Jedi, Dooku began to train young Qui-Gon Jinn.''\\
  \cline{2-0}
  excerpts, $M$  & \textbf{Doctor} \\
  & ``Amy Wong and Leela drag Fry and Bender to the gym and Doctor Zoidberg tags along.''\\ 
  & ``Elsewhere on Deep Space 9, Doctor Bashir is called to the wardroom''\\
     \cline{2-0}
     & \textbf{Kirk} \\
     &``Zapp Brannigan is holding a court martial in the Planet Express Ship, ...'' \\
     &`` When Riker explains that they came from the starship Enterprise, Scott reacts under...''
\end{tabular}
  \label{tab:scifi-explain}

\end{table*}


We specifically investigate three case studies from our dataset: a Star Wars villain named ``Dooku'', a Star Trek hero ``Kirk'', and the main character of Doctor Who, the ``Doctor.'' We observe that Dooku is ultimately placed into a group of villains (including terms such as ``exploits,'' ``confrontation,'' and ``fighters''), Kirk is near other Star Trek captains and characteristics of his character (including  words such as ``Janeway,'' ``beams,'' and ``flirting''), and the Doctor is placed into a Doctor Who neighborhood (including other names from the show such as ``Donna,'' ``Rory,'' and ``Gallifrey'').

We ask, ``Which samples moved each word to their final neighborhood?'' and ``Which samples would push them out?'' Applying CLIF to $D_{SCI}$, we surface the most influential plot summaries for each word embedding and present a subset of instances in Table \ref{tab:scifi-explain}.

We observe that samples returned by CLIF intuitively match the neighbors for each word in our case study. For Dooku, CLIF returns reinforcing plot summaries involving Dooku as a villain (``Dooku as a place holder Sith apprentice and ally in the war,''), while plot summaries that would move Dooku \textit{away} from his final position include flashback plots before his character turned evil (``now ... Jinn,''). The Doctor's reinforcing plot summaries include Doctor Who episodes, while other uses of the word ``doctor'' would move the word embedding away from its Doctor Who neighborhood (e.g., references to doctors in other shows). Finally, we observe that original Star Trek episodes reinforce Kirk's association with Star Trek concepts, while references to Kirk from other shows, such as Futurama, serve to move Kirk's embedding away due to the abnormal use of ``Kirk'' in novel contexts (e.g., with new characters and locations).

\subsection{Model Bias: Wiki Neutrality Corpus} 
\label{subsec:wnc-exp}
We have thus far shown that  our approach successfully discovers influential samples for a synthetic MOG dataset and a dataset of plot summaries, each of which is we can manually review. Next, we evaluate CLIF on a relatively large dataset--large enough to prevent full human annotation. We leverage a larger dataset of Wikipedia edits from Pryzant \textit{et al.}~\citep{pryzant2019automatically} containing 180,000 sentences pairs, each consisting of a point-of-view biased sentence and its neutralized counterpart. Due to this dataset's significantly increased size, this experiment provides us with a domain which is much more representative of a real-world machine learning deployment scenario for de-biasing. 

In order to quantify bias with an objective function, we leverage the Word Embedding Association Test (WEAT) \citep{caliskan2017semantics}.
For each set of ``target'' words (e.g., \textit{male}) and ``attribute'' words (e.g., \textit{math}) which comprise a WEAT, the WEAT score indicates the effect size for a bias in the word embeddings\footnote{A formal definition is included in the appendix.}.
The tests for this dataset involve four relevant WEATs from prior work: gender-career ($WEAT_C$), gender-math ($WEAT_M$), gender-art ($WEAT_A$), and race-pleasantness ($WEAT_R$). 

CLIF for a WEAT objective is shown in Equation \ref{eqn:weat-influence-function}. The WEAT loss function uses the absolute value of the effect size as the test loss value, and asks, ``Which samples are most responsible for high amounts of bias?'' Explanatory examples and results are presented here for the set of all point-of-view biased samples, $D_B$, and in the appendix for the set of neutralized samples, $D_N$.

\begin{table*}
  \caption{Examples of Influential Samples for Word Embedding Association Test Scores in $D_B$}

  \begin{tabular}{lcll}
    WEAT&Set&Example&Cause\\
    \hline
    Career&$A$&``The vice president has an office in the building but his primary office& Aligns male \\
    && is next door in the Eisenhower executive office building.''& and business.\\
    
    Career&$M$&``Classical violinist Joshua Bell joins bluegrass musicians Sam Bush and & Breaks alignment of \\
    &&Mike Marshall and the versatile Meyer on the album.''& male and career. \\
    
    \hline
    Math&$A$&``Maxwell's $\langle NUM \rangle$ formulation was in terms of $\langle NUM \rangle$ & Aligns male\\
    && equations in $\langle NUM \rangle$ variables although he later attempted & and math. \\
    && a quaternion formulation that proved awkward''& \\
    
    Math&$M$&``Most programming languages are designed around computation & Aligns female \\ 
    &&whereas Alice is designed around storytelling and thus has greater &and math. \\
    &&appeal to female students.''& \\
    \hline
    Art&$A$&``Gentry claim that his critics are not able to supply actual& Aligns male\\
    &&scientific evidence to combat his work''& and science.\\
    

    Art&$M$&``However his subsequent performance received almost universal acclaim.''& Aligns male and art. \\

    \hline
    
    Race&$A$&``Afterward, Mario, Peach, and the others then begin their well &Aligns EA names \\
    &&deserved vacation&and pleasant\\

    Race&$M$&``President Barack Obama returned from vacation sporting &Aligns AA names \\
    &&Oliver Peoples sunglasses&and pleasant\\
    \hline
\end{tabular}
  \label{tab:bias-examples}

\end{table*}

\begin{align}
    \label{eqn:weat-influence-function}
    \mathcal{L}(WEAT, z_{tr}) = -\nabla_\theta WEAT^T H_{\theta}^{-1} \nabla_\theta L_{SG}(z_{tr})
\end{align}


To establish the bias inherent in the datasets, the results of the four WEATs 
are given in Table \ref{tab:debias-results} as ``Original.'' 
We leverage CLIF for WEAT (Equation \ref{eqn:weat-influence-function}) to identify sources of bias, as shown in Table \ref{tab:bias-examples}. We note that many examples which are best for mitigating the bias inherent in the word embeddings are not necessarily neutral but are instead biased in an orthogonal or antilinear direction to the existing bias, such as aligning \textit{male} with \textit{art}. This finding is discussed further in Section \ref{sec:discussion}. The examples our method returns for amplifying or mitigating bias demonstrate that our method can open the black box of representation learning and explain the genesis word embedding properties.

\subsubsection{Case Study: Hidden Influential Concepts}
\label{subsubsec:bias-case-study}
CLIF enables us to find seemingly unrelated concepts that are highly influential to model bias. For example, investigating the samples which are most influential for the $WEAT_R$ over $D_N$, reveals that two of the most influential samples appear completely unrelated to the $WEAT_R$:
\begin{flushitemize}
    \item``In Jewish self-hatred, a person exhibits a strong shame or hatred of their Jewish identity, of other Jews, or of the Jewish religion.'' \\
    \item ``Islamophobia is the irrational fear and or hatred of Islam, Muslims, or Islamic culture.''
\end{flushitemize}
We can see that there are no words from the WEAT target set for \textit{race}, or words from the \textit{pleasant} set, but instead the samples align a \textit{religion} concept with \textit{unpleasant}. This finding is important, as it suggests that even a manual review of datasets may be inadequate -- unwanted biases may be lurking in non-obvious correlations that are only revealed through a more quantitative approach.

To understand these non-obvious examples, we construct a WEAT with only one set of attribute concepts, comparing the target concepts (\textit{names}) in $WEAT_R$ to the \textit{religious} terms above. We find that \textit{African-American names} have twice the mean cosine-similarity to \textit{religious} terms (0.12 vs 0.06) compared to \textit{European-American}, and the one-sided WEAT score is $-1.11$. This alignment sheds light on the curious set of influential examples. Rather than finding examples where \textit{unpleasant} terms coincide with \textit{African-American names}, we found samples in which concepts that are strongly aligned with \textit{African-American names} (i.e., \textit{religion}) are aligned with \textit{unpleasant} terms. Not only is our approach crucial for revealing such findings, but we also present avenues to mitigate these biases (Section \ref{subsec:bias-mitigation}).

\subsection{Bias Removal or Reinforcement}
\label{subsec:bias-mitigation}

In our final experiment, we empirically validate our approach to explainability by using the results of our influence estimation to augment properties of word embeddings. With an ordered set of all influential examples, we separate the samples into two sets which have the greatest effect on bias: training examples that amplify bias, $A$, and training examples that mitigate bias, $M$. We assert that we have found influential samples if we are able to reduce bias according to WEAT by removing the effects of the bias-amplifiers ($A$) or by reinforcing the effects of bias-mitigators ($M$). We remove the effects of $A$ by calculating training losses for each sample, $a \in A$, and then taking gradient steps in the opposite direction. Similarly, we reinforce the effects of $M$ by fine-tuning over $M$. Finally, we experiment with the combination of both removing the effects of $A$ and reinforcing the effects of $M$ in tandem. Set sizes for $A$ and $M$ are given in the appendix.

In these experiments, we compare our fine-tuning debiasing to gender-dimension debiasing from prior work \citep{bolukbasi2016man}. We test fine-tune debiasing both with and without gender-dimension debiasing, showing that it always yields word embeddings with a weaker biased effect (i.e., more neutral embeddings). 
Results for our experiments are given in Table \ref{tab:debias-results}, where a zero-effect size indicates neutral embeddings according to WEAT.

We find that augmenting word embeddings with training examples in $A$ and $M$ reduces and even removes bias as measured by WEAT. We present an auxiliary experiment in the appendix in which we exchange sets $A$ and $M$ and see that they further polarize the data and \textit{increase} bias. Taken together, these findings further demonstrate that our approach finds samples of importance (i.e. strong cross-loss influence) and that fine-tuning with such samples meaningfully affects dataset bias according to WEAT.

Finally, we compare the method from \citet{bolukbasi2016man} to ours on both $D_B$ and $D_N$. In Table \ref{tab:debias-results}, we find that our approach is effective at reducing high gender and racial bias. On the other hand, the method of \citet{bolukbasi2016man} performs very poorly for racial bias. This poor result can be explained by a reliance on an assumed ``bias-vector.'' Unfortunately, if the bias' magnitude is not large enough or if the bias-vector is inaccurate, applying the method of \citet{bolukbasi2016man} can increase bias. This effect is visible in the  \textit{Gender Career} test on $D_N$ in Table \ref{tab:debias-results}.

As our approach does not make assumptions about dimensions of bias vectors, our method may be applied even if bias is already low. We can therefore apply prior work~\citep{bolukbasi2016man} first and then apply our fine-tuning on top, an approach shown in the ``Ours (post \citep{bolukbasi2016man})'' row of Table \ref{tab:debias-results}. This combination almost always yields the most neutral results. Ultimately our final experiment has shown that our approach to explainability is quantitatively verified on a large dataset of real-world language data, enabling insight into sources of bias and enabling us to even remove unwanted model bias without the tenuous assumptions of prior work. 
\begin{table*}
\small
\caption{WEAT Effect Sizes Across $D_B$ and $D_N$. Zero Means Unbiased. Bold Highlights Least Bias. \\ }
\begin{tabular}{lcccc|cccc}
     &\multicolumn{4}{c}{$D_B$}&\multicolumn{4}{c}{$D_N$}\\
     Method&Gender&Gender&Gender&Race&Gender&Gender&Gender&Race\\
     &Career&Math&Arts&Pleasant&Career&Math&Arts&Pleasant\\
     \hline
     \hline

     Original&0.48&0.52&-0.43&0.53&-0.05&0.66&-0.82&0.55\\
     \citet{bolukbasi2016man}&0.008&0.12&-0.04&0.53&-0.29&0.16&0.09&0.71\\
     \hline
     Ours&0.01&0.10&\textbf{0.01}&\textbf{0.02}&\textbf{-0.01}&-0.02&-0.11&-0.07\\
     Ours (post
     ~\citet{bolukbasi2016man})
     &\textbf{0.000}&\textbf{0.009}&-0.05&0.07&-0.07&\textbf{-0.01}&\textbf{-0.01}&\textbf{-0.03}\\
     \hline
     \vspace{0.02cm}

\end{tabular}
\label{tab:debias-results}

\end{table*}

\section{DISCUSSION}
\label{sec:discussion}
Our novel derivation of CLIF brings newfound explainability to deep network representations through identification of relevant training examples. Our method permits insight into deep network representations, even with unsupervised and self-supervised training objectives, and offers an approach to mitigating bias by fine-tuning a trained model with discovered samples. Our experimental results in Section \ref{sec:experiments} demonstrate the powerful potential of our method to both explain and manipulate deep network representations. 

In our synthetic data domain (Section \ref{subsec:synthetic-exp}), we evaluated the predicted and actual influence of all samples in the dataset using both matched and mismatched objectives. When compared to ground-truth influence, we found that mismatched objectives with CLIF yielded higher average correlation (65.5\%$>0.8$) than influence functions with matched-objective settings (58\%$<0.8$). Through this experiment, we empirically validated our approach in an example deployment scenario of unsupervised pre-training with multiple classes, before moving into the word embedding domains.

Applying CLIF to explaining properties of science-fiction concepts, we provide another case study of our approach applied to explaining nearest-neighbors. After examining the nearest-neighbors of different word embeddings (Dooku, Kirk, and Doctor), we find that CLIF surfaces highly relevant plot summaries for each word both as reinforcing examples (e.g., original Star Trek or Doctor Who episodes) or removing examples (e.g., Futurama episodes).

In our bias explainability domain (Section \ref{subsec:wnc-exp}), we discovered that samples which mitigate bias are often ones which would encourage greater diversity in the target sets. For example, we see in Table \ref{tab:bias-examples} that one sample which lowers the $WEAT_C$ score for $D_B$ is one which aligns \textit{male} with \textit{music}. While $WEAT_C$ tests for relative career-family alignment, encouraging \textit{male} to move closer to \textit{music} is also a viable strategy for reducing the strength of the \textit{male-career} association relative to the \textit{male-family} or \textit{female-career} associations. Similar approaches to reducing the $WEAT$ scores can be seen throughout Table \ref{tab:bias-examples} and the appendix, where simply aligning \textit{male} and \textit{female} concepts with more diverse concepts can have the effect of lowering overall bias. We find that these diversification trends apply unless the $WEAT$ effect is already sufficiently close to zero, as the $WEAT_C$ in $D_N$.

Finally, we have shown that our work improves the state of the art in bias-dimension neutralization through complementary fine-tuning on relevant training data. 
As prior work is founded on the assumption that the principal component of a set of biased words is correlated with a bias dimension, we find that their approach \textit{increases} bias if applied to a set of already-neutralized embeddings, as this principal component assumption is then no longer valid. However, with enough of a ``gender'' component, prior work can significantly lower bias, and our approach can then continue to reduce bias that is missed by prior work, nearly perfectly neutralizing WEAT effects. As prior work only includes a very small list of racial words to establish a ``race'' dimension, the method underperforms on the $WEAT_R$ test. Fine-tuning with CLIF circumvents this issue, permitting modification of learned embeddings with influential samples and having more substantial effect on model bias.



\section{LIMITATIONS}
\label{sec:limitations}
Our work is predicated on two core assumptions. First, we assume that downstream task performance can be predicted by first-order interactions with a training objective. For example, in our WEAT and Skip-gram embedding examples, our work assumes that WEAT effects can be explained directly by the Skip-gram losses in our training data. However, as we showed in our case study on hidden concepts (Section \ref{subsubsec:bias-case-study}) such assumptions are not always valid. In cases where downstream task performance (e.g. a WEAT effect) is explained by second-order relationships, our method will yield seemingly unrelated training data (e.g. sequences about \textit{religious} concepts).

The second core assumption within our work is that we have access to all training data used for the model. Influence calculation for training data requires access to the training data, in order to calculate each sample's effect on some cross-loss objective. In private federated learning setups, continual learning setups, or other paradigms where we may not retain access to static training data, then our method will not be able to highlight historically relevant samples-- we can only consider samples that we can directly process.

\section{SOCIETAL IMPACT}
\label{sec:societal-impact}
Deep learning is prevalent in the real world, and embedding models are foundational to the deployment of deep learning at-scale. Whether as video-encoders, word-embedding models, speech-encoders, or others, embedding models enable the success of on-device machine learning and provide the basis for modern machine learning research and deployment. Furthermore, prior research has already shown that many deployments of machine learning research contain inherent biases \citep{silva2021towards,bolukbasi2016man,lum2016predict}. As such, it is imperative that the research community develop and evaluate a diverse set of tools for discovering and eliminating bias. Our work contributes one such tool, which we demonstrate may be used in conjunction with prior work or for independent dataset interrogation.

Crucially, our work relies on a definition of bias and an objective function to represent that definition. Proper application of our approach to de-biasing or fairness will require careful definition of such objectives. Recent work \citep{blodgett2020survey} has called bias research into question for poor motivation, unclear definition, and a lack of grounding. In our work, ``bias'' refers to a mismatch in alignment between neutral words, such as ``microscope,'' with gendered words, such as ``brother'' and ``sister.'' The WEAT \citep{caliskan2017semantics} was adapted from well-established psychology literature on the Implicit Association Test \citep{greenwald1998measuring} as a means of estimating inequality in word embedding cosine similarities. We specifically investigate WEAT effects in our work, and are therefore investigating a mismatch in cosine similarities of embeddings when we discuss bias. Implications for this bias are explored further in the original WEAT work, specifically that language models may ``acquire historical cultural associations, some of which may be objectionable,''\citep{caliskan2017semantics} and that these associations may be propagated or misused in downstream applications. Our work explicitly offers explainability into the WEAT and its effects, and we make no further claims with respect to its definition of bias or its implications. To apply our approach to other measures of bias and fairness would require engineering and evaluating a new objective.

\section{CONCLUSION}
\label{sec:conclusion}
We have presented an extension to influence functions for estimating sample importance in deep networks with arbitrary train and test objectives. Our extension allows us to explain deep network representations by revealing which training samples are the most relevant to final test performance, even for tests which are unrelated to the training task, such as bias tests. We demonstrated this approach by explaining various deep network representations, as well as by fine-tuning Word2Vec models to alter the model's inherent biases.
Our approach affords greater explainability across a variety of problems and makes the influence function more broadly available to researchers in machine learning as a tool for model and dataset interrogation.

\section*{Acknowledgments}
This work was supported by NASA Early Career Fellowship grant 80HQTR19NOA01-19ECF-B1.

\bibliographystyle{plainnat}
\bibliography{sample-base}

\clearpage
\include{supplementary_material}

\end{document}

%% file: supplementary_material.tex
\onecolumn
\aistatstitle{Supplementary Material for: Cross-Loss Influence Functions to Explain Deep Network Representations}
\author{Anonymized for Peer Review}






\appendix
\section{CLIF Algorithmic Walkthrough}
Here we present a pseudocode block for the CLIF calculation in our work. Calculation of gradients is done using PyTorch (e.g. list(grad(loss\_val, list(model.parameters()), create\_graph=True))). Our code is available at https://github.com/CORE-Robotics-Lab/Cross\_Loss\_Influence\_Functions
\begin{algorithm*}
\caption{CLIF Calculation}
\label{alg:training_loop}
\begin{algorithmic}[1]
\STATE {\bfseries Given:} Test objective, $\mathcal{L}_{Te}$, Twice-differentiable training objective, $\mathcal{L}_{Tr}$
\STATE{\bfseries Given:} Trained model parameters $\theta$, Dataset $D$
\STATE{\bfseries Given:} Embeddings to explain $\epsilon$
\STATE{\bfseries Initialize:} Influential samples $E \leftarrow \emptyset$
\FOR{$d \in D$}
\STATE $z_d \leftarrow \nabla_\theta\mathcal{L}_{Te}(\theta, \epsilon)$ // Calculate test-size effect on $\epsilon$ under $\theta$
\STATE $l_d \leftarrow \nabla_\theta \mathcal{L}_{Tr}(\theta, d)$ // Calculate training impact of $d$ on $\theta$
\STATE $h_d \leftarrow \nabla_\theta (l_d*z_d)$ // Compute Hessian of $l_d$ with respect to $z_d$ under $\theta$
\STATE $E \leftarrow E \cup (l_d \cdot h_d)$ // Store magnitude of gradient effects on $h_d$ with respect to $l_d$ under $\theta$
\ENDFOR
\RETURN sorted(E) // Return ordered list of influential samples
\end{algorithmic}
\end{algorithm*}

\section{WEAT Loss}
In order to quantify bias with an objective function, we leverage the Word Embedding Association Test (WEAT) \citep{caliskan2017semantics}.
For each set of ``target'' words (e.g., \textit{male}) and ``attribute'' words (e.g., \textit{math}) which comprise a WEAT, we take every word $\vec{w}$ in target sets $X$ and $Y$ and compute its average cosine similarity with each set of attributes, $A$ and $B$. With $cs$ as the cosine-similarity function, the score is given in Equation \ref{eqn:weat-score}, and the $WEAT$ effect is given in Equation \ref{eqn:weat-effect}.
\begin{align}
    \label{eqn:weat-score}
    s(\vec{w}, A, B) = \displaystyle \mathop{\mathbb{E}}_{\vec{a} \sim A}[cs(\vec{w}, \vec{a})] - \displaystyle \mathop{\mathbb{E}}_{\vec{b} \sim B}[cs(\vec{w}, \vec{b})]
\end{align}
\begin{align}
    \label{eqn:weat-effect} 
     WEAT = \frac{\displaystyle \mathop{\mathbb{E}}_{\vec{w} \sim X}[s(\vec{w}, A, B)] - \displaystyle \mathop{\mathbb{E}}_{\vec{w} \sim Y}[s(\vec{w}, A, B)]}{\displaystyle \mathop{\mathbb{E}}_{\vec{w} \sim X \cup Y}[\sigma(s(\vec{w}, A, B))^2]}
\end{align}

\begin{align}
    \label{eqn:weat-influence-function-appendix}
    \mathcal{L}(WEAT, z_{tr}) = -\nabla_\theta WEAT^T H_{\theta}^{-1} \nabla_\theta L_{SG}(z_{tr})
\end{align}

The tests for this dataset involve four relevant WEATs from prior work: gender-career ($WEAT_C$), gender-math ($WEAT_M$), gender-art ($WEAT_A$), and race-pleasantness ($WEAT_R$). 

Our cross-loss influence function for a WEAT objective is shown in Equation \ref{eqn:weat-influence-function-appendix}. The WEAT loss function uses the absolute value of the effect size as the test loss value.

\subsection{WEAT Derivative}

The WEAT objective consists of a set of cosine similarity comparisons, computing the difference in mean cosine-similarity between two sets of words (the target sets) with a third set (the attribute set). To compute the WEAT derivative analytically, one must compute the derivative of the cosine similarity between two words (given in Equation \ref{eqn:cosine-derivative}), and apply the product and chain rules to the resulting derivative equations.

\begin{equation}
\label{eqn:cosine-derivative}
    \frac{\partial cos(a, b)}{\partial a_i} = \frac{b_i}{|b|\cdot|a|}-cos(a,b)\frac{a_i}{|a|^2}
\end{equation}

\section{WEAT Examples}
Examples of the WEATs that we use are given in Table \ref{tab:WEAT-examples}. Note that some tests, such as \textit{Math} and \textit{Art} overlap very heavily, meaning that correcting one score often involves correcting both.

\begin{table*}[b]
\centering
\small
  \begin{tabular}{llllll}
\hline
    Test&Symbol&$X$&$Y$&$A$&$B$\\
\hline
    Career&$WEAT_C$&john, greg, jeff&amy, lisa, donna&executive, office, salary&home, wedding, family\\
    Math&$WEAT_M$&male, boy, he&female, girl, she&math, geometry, numbers&art, novel, drama\\
    Art&$WEAT_A$&son, his, uncle&daughter, hers, aunt&poetry, dance, novel&science, physics, nasa\\
    Race&$WEAT_R$&josh, katie, jack&theo, jasmine, leroy&freedom, honor, love&crash, ugly, hatred\\
\hline
\end{tabular}
  \caption{Word Embedding Association Test set examples. Tests are drawn from the original WEAT work~\citep{caliskan2017semantics}.}
  \label{tab:WEAT-examples}
\end{table*}

\section{Over-Biasing Experiments}
\label{sec:over-bias}
We conduct an auxiliary experiment to validate the claim that our approach has found samples which are relevant to the model's inherent bias. Rather than attempting to undo or neutralize inherent bias, we instead reverse the process described in the main paper and instead \textit{amplify} the inherent biases.

Ordinarily, we mitigate bias by reinforcing the effects of ``mitigating'' samples, $M$, and undoing the effects of ``amplifying'' samples $A$. In this experiment, we swap sets $M$ and $A$, polarizing the models as much as possible. Our results are given in Tables \ref{tab:over-bias} \& \ref{tab:over-neutral}. Curiously, the $WEAT_R$ effect flips for both datasets, $D_B$ and $D_N$. Even so, the overall bias in each dataset is clearly exacerbated, as WEAT effects are nearly perfectly polarized for most tests. Even the near-neutral $WEAT_C$ on $D_N$ has been increased substantially. While we would never want to apply such dataset augmentations in the wild, the ability of our discovered samples to so greatly amplify inherent dataset biases provides addition empirical support for our approach.
\begin{table}[]
    \centering
    \small
    \begin{tabular}{l|cccc}
    Method&Gender&Gender&Gender&Race\\
    &Career&Math&Arts&Pleasant\\
  \hline
    Original&0.48&0.52&-0.43&0.53\\
    Over-Biased&0.94&1.61&-1.63&-1.72\\
    \hline
    \end{tabular}
    \caption{$D_B$ WEAT effects after over-biasing}
    \label{tab:over-bias}
\end{table}

\begin{table}[]
    \centering
        \small

    \begin{tabular}{l|cccc}
    Method&Gender&Gender&Gender&Race\\
    &Career&Math&Arts&Pleasant\\
  \hline
    Original&-0.05&0.66&-0.82&0.55\\
    Over-Biased&-0.69&1.85&-1.58&-1.61\\
        \hline

    \end{tabular}
    \caption{$D_N$ WEAT effects after over-biasing}
    \label{tab:over-neutral}
\end{table}

\section{Bias Experiments on $D_N$}
\begin{figure*}[t]
\centering
\begin{subfigure}[b]{\textwidth}
\centering
        \includegraphics[width=0.55\textwidth, height = 0.5cm]{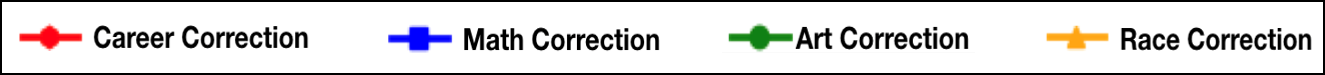}
\end{subfigure}
    \begin{subfigure}[b]{0.23\textwidth}
        \includegraphics[width=0.95\textwidth]{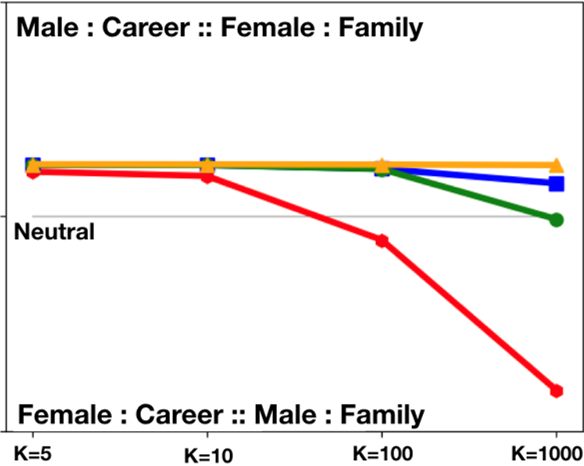}
        \caption{Career Score, $WEAT_C$}
        \label{fig:biased-weat-c-results}
    \end{subfigure}
    ~~
    \begin{subfigure}[b]{0.23\textwidth}
        \includegraphics[width=0.95\textwidth]{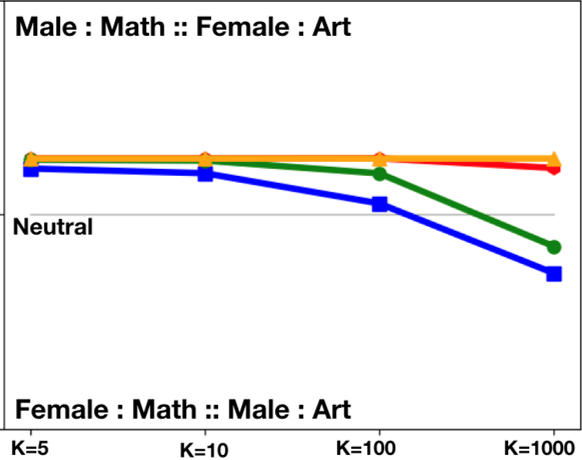}
        \caption{Math Score, $WEAT_M$}
        \label{fig:biased-weat-m-results}
    \end{subfigure}
    ~~
    \begin{subfigure}[b]{0.23\textwidth}
        \includegraphics[width=0.95\textwidth]{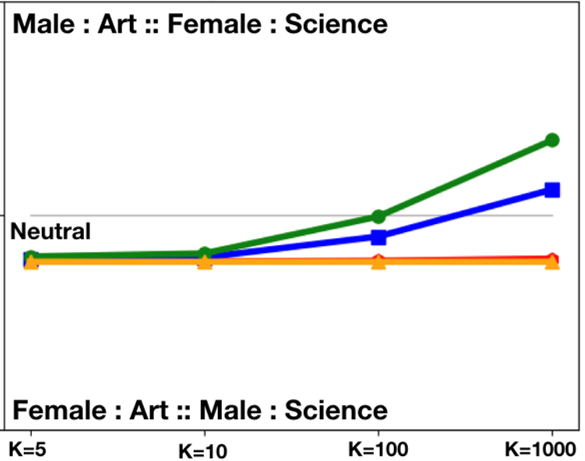}
        \caption{Art Score, $WEAT_A$}
        \label{fig:biased-weat-aresults}
    \end{subfigure}
    ~~
    \begin{subfigure}[b]{0.23\textwidth}
        \includegraphics[width=0.95\textwidth]{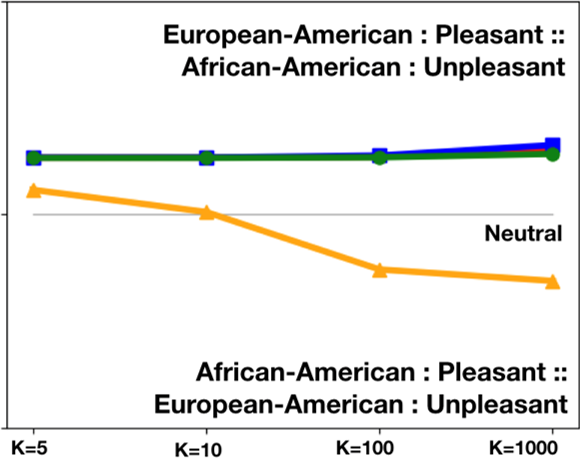}
        \caption{Race score, $WEAT_R$}
        \label{fig:biased-weat-rresults}
    \end{subfigure}
\caption{For the biased dataset, $D_B$, we observe the effects on WEAT scores as we modify the augment the word embedding model with samples discovered from our approach.}
\label{fig:bias-results}
\end{figure*}

\begin{figure*}[t]
\centering
\begin{subfigure}[b]{\textwidth}
\centering
        \includegraphics[width=0.55\textwidth, height = 0.5cm]{fig/best_biased_legend.png}
\end{subfigure}
    \begin{subfigure}[b]{0.23\textwidth}
        \centering
        \includegraphics[width=0.95\textwidth]{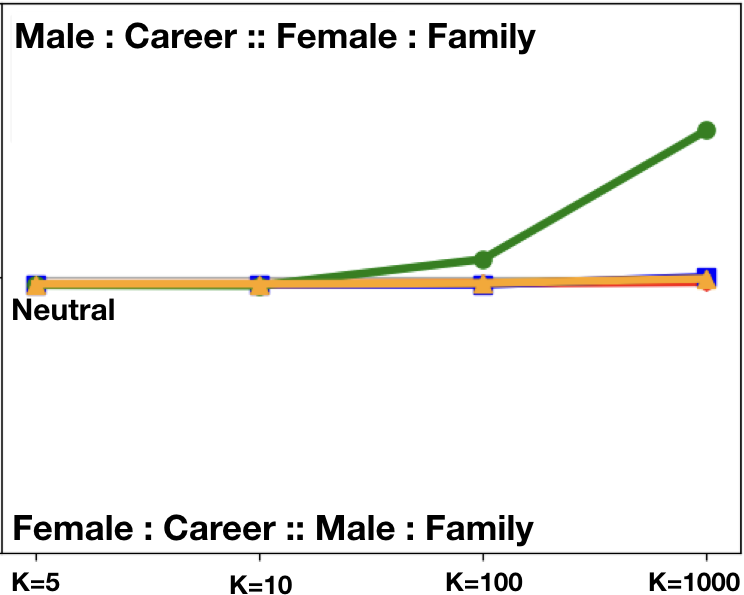}
        \caption{Career score, $WEAT_C$}
        \label{fig:neutral-weat-c-results}
    \end{subfigure}
    ~~
    \begin{subfigure}[b]{0.23\textwidth}
        \centering
        \includegraphics[width=0.95\textwidth]{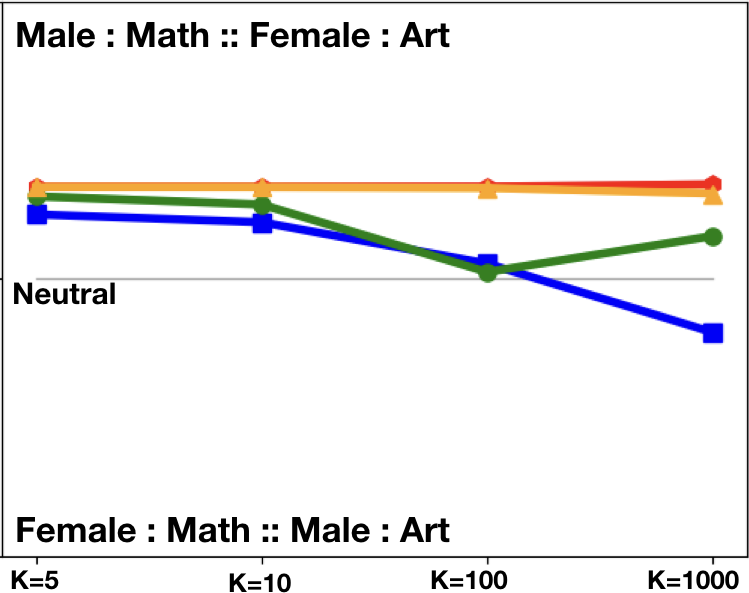}
        \caption{Math score, $WEAT_M$}
        \label{fig:neutral-weat-m-results}
    \end{subfigure}
    ~~
    \begin{subfigure}[b]{0.23\textwidth}
        \centering
        \includegraphics[width=0.95\textwidth]{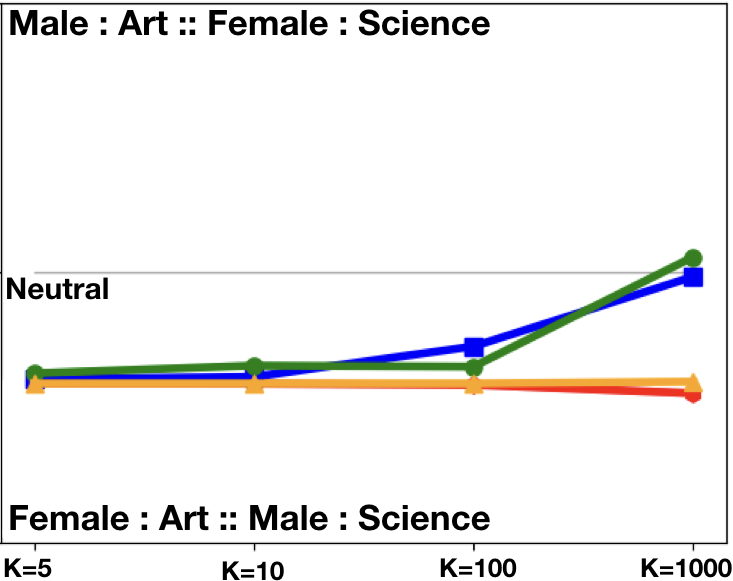}
        \caption{Art score, $WEAT_A$}
        \label{fig:neutral-weat-aresults}
    \end{subfigure}
    ~~
    \begin{subfigure}[b]{0.23\textwidth}
    \centering
        \includegraphics[width=0.95\textwidth]{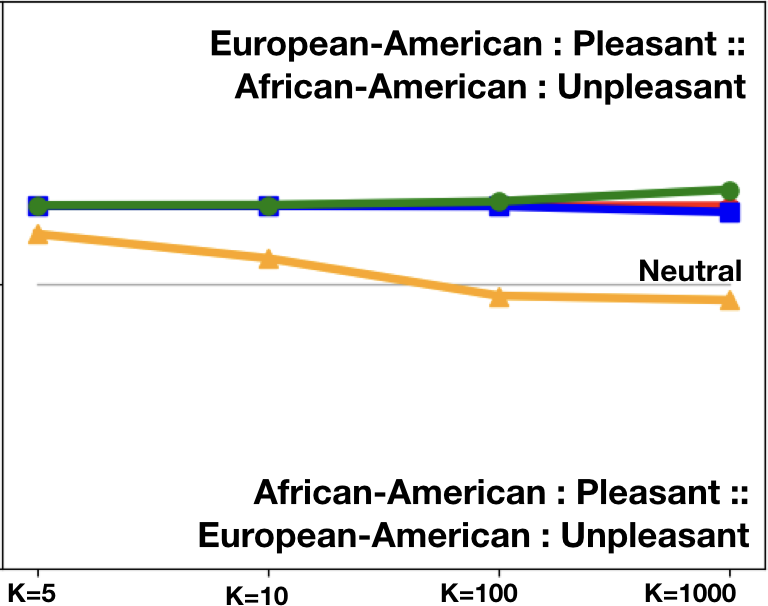}
        \caption{Race score, $WEAT_R$}
        \label{fig:neutral-weat-rresults}
    \end{subfigure}
\caption{For the neutralized dataset, $D_N$, we observe the effects on WEAT scores as we modify the augment the word embedding model with samples discovered from our approach.}
\label{fig:neutral-results}
\end{figure*}

In this section, we include results which further support our central hypotheses: proof of bias mitigation through amplifying and mitigating samples across four WEATs, given in Figures \ref{fig:bias-results} \& \ref{fig:neutral-results}, and a set of samples which are among the highest in influence on the WEAT scores, given in Table \ref{tab:neutral-examples}. In each subfigure, we show the score on the corresponding WEAT for four different sets of updated word embeddings: corrections applied to neutralize the score on $WEAT_C$ (Career Correction), $WEAT_M$ (Math Correction), $WEAT_A$ (Art Correction), and $WEAT_R$ (Race Correction). We expect to see that corrections for each test effectively undo the discovered bias, and that the gender-aligned tests ($WEAT_C$, $WEAT_M$, and $WEAT_A$) move together. 

We see an interesting case in $D_N$ in which initial bias is already very low ($WEAT_C$). The initial bias test scores very near zero at $-0.05$, and resulting training examples are all quite neutral. Even when applied for hundreds or thousands of iterations, the score on the bias test barely shifts upwards, gently approach zero but never significantly over-correcting (Figure \ref{fig:neutral-weat-c-results}).

To our surprise, we find that $D_N$ contains higher baseline bias than $D_B$, despite containing copies of sentences in $D_B$ which are explicitly neutralized to have no point of view. While counter-intuitive at first, the result is reasonable as we examine more of the data which is responsible. For instance, the sentence ``Harris left NASA in April $\langle NUM \rangle$ but he continued in research,'' does not contain bias which can be neutralized, the content simply encourages alignment between \textit{male} and \textit{science}. 


As shown in Table \ref{tab:neutral-examples} and in examples from the main paper, we find that some of the strongest examples for each dataset are the same, even after neutralization. In particular, the examples which affect the $WEAT_M$ score are often the same for $D_B$ and $D_N$. Often times, these examples are reflective of the same \textit{male} and \textit{science} example given above: content itself is inherently aligning genders and attributes, even without being bigoted or sexist in a conventional sense. The finding that perspective-neutral content is responsible for bias again reinforces the need for approaches which can both explain and mitigate bias in word embeddings, as we are not simply able to screen out all content which introduces bias.

We find that there is a careful balance to strike between under- and over-correcting, a challenge shown in Figure \ref{fig:biased-weat-c-results}. While updating for 10 -- 100 iterations might result in completely neutral gender-career word embeddings, applying the updates 1000 times ends up significantly skewing the embeddings in the other direction, over-aligning \textit{male} with \textit{family} and \textit{female} with \textit{career}. 

While initially surprising, this result makes more sense as we begin to examine some of the members of sets $A$ and $M$ (Table \ref{tab:neutral-examples}), where we see that training examples which reduce bias are rarely neutral. Instead, when bias has a sufficiently high absolute value, the most helpful samples for correction are those which will over-correct when over-applied. 

We find that the fine-tuning with correction sets for related effects moves related biases. For example, applying the Art Correction neutralizes the art bias, but also neutralizes career bias, as shown in Figure \ref{fig:biased-weat-c-results}. As we can see in Table \ref{tab:neutral-examples}, one of the most influential samples to mitigate bias in $WEAT_M$ is one which aligns \textit{female} and \textit{math}, which has a side-effect of neutralizing $WEAT_A$.


\begin{table*}
\centering
\begin{tabular}{lcll}
\hline
    WEAT&Set&Example&Cause\\
\hline
    Career&$A$&``It has elected people to public office.''& Neutral \\
    
    Career&$M$&``They had a long marriage with no children.''& Neutral \\
\hline
    
    Math&$A$&``Maxwell's $\langle NUM \rangle$ formulation was in terms of $\langle NUM \rangle$  & Aligns male  \\
    &&equations in $\langle NUM \rangle$ variables although he later &and math\\ 
    && attempted a quaternion formulation''&\\
    
    Math&$M$&``Most programming languages are designed around computation & Aligns female \\
    && whereas Alice is designed around storytelling and thus is claimed & and math  \\ 
    && to have greater appeal to female students.''&\\
\hline
    Art&$A$&``With her dance moves she is best known for her spins & Aligns female \\ 
    && and her hip hop moves.''&and art \\
    
    Art&$M$&``His first album after his $\langle NUM \rangle$ farewell concerts was dance 1981.''& Aligns male \\
    &&&and art \\

\hline

    Race&$A$&``It is a popular place for camping and recreational vacation.''&Aligns pleasant \\
    &&&topics \\
    
    Race&$M$&``Nate Holden performing arts center, the center at $\langle NUM \rangle$ & Aligns African- \\
    && 18 west Washington boulevard is the home of the & American and art \\
    && Ebony Repertory theater company.''& \\
    
\end{tabular}
  \caption{Examples of influential samples for Word Embedding Association Test scores in $D_N$. Many samples are important in both versions of the WNC, indicating their importance in bias propagation.}
  \label{tab:neutral-examples}
\end{table*}

\section{Amplifying and Mitigating Set Sizes}
Amplifying set $A$ and mitigating set $M$ sizes for each dataset ($D_B$ and $D_N$) and correction set ($WEAT_C$ -- Career Correction, $WEAT_M$ -- Math Correction, $WEAT_A$ -- Art Correction, and $WEAT_R$ -- Race Correction) are given in Table \ref{tab:set-sizes}.
\begin{table}[h]
\small
\centering
  \begin{tabular}{lll}
    \hline
    Correction Set&$D_B$&$D_N$\\
    \hline
    Gender-Career $A$&0&5\\
    Gender-Career $M$&100&0\\
    Gender-Math $A$&0&0\\
    Gender-Math $M$&100&100\\
    Gender-Art $A$&0&1000\\

    Gender-Art $M$&100&0\\
    Race-Pleasantness$A$&0&0\\
    Race-Pleasantness$M$&5&5\\
    \hline
\end{tabular}
  \caption{Amplifying and Mitigating set sizes for $D_B$ and $D_N$.}
  \label{tab:set-sizes}
\end{table}

\section{Dataset Tokenization}
Tokenization is a crucial piece to any language modeling project. In this section, we outline and justify our tokenization choices for the two datasets we considered in this work.

\paragraph{Science Fiction}  The goal of our experiments with $D_{SCI}$ is to cluster meaningful concepts (e.g., heroes vs. villains, or Star Trek vs. Star Wars). To that end, we remove common stopwords and all words which are not represented at least five times in the corpus, leaving us with a vocabulary of 15,282 tokens. Using words which are represented at least five times means that we will have a denser corpus of commonly used tokens. Similarly, stopwords are common in all plot summaries, but they do not contain information which we wish to capture.

\paragraph{WNC} The goals of our experiments with $D_B$ and $D_N$ are twofold: explainability around gender and racial bias, and augmentation of discovered biases. Because this set of experiments involves testing for gender bias specifically around pronouns (which are common stopwords), we do not remove stopwords for this dataset. Instead, we use the set of tokens generated in prior work, giving us a vocabulary of 25,355 tokens. Using a pre-made tokenization allows us to improve the reproducibility of our work and helps us to avoid injecting our own bias into the words that we search for. We use the same set of tokens for each dataset.

\section{Word Embedding Model Training}
In this section, we briefly cover hyperparameters which would be necessary to reproduce our word embedding models. For both approaches, negatives samples are drawn randomly from the entire training corpus ($D_{SCI}$, $D_B$, or $D_N$), and words are sampled according to their prevalence in the dataset.

\paragraph{Science Fiction} We train an embedding model over $D_{SCI}$ for 100 epochs with a context window of three, five negative samples per word-context pair, and an embedding dimension of 100. Training for so long allows for the embeddings to form dense clusters with related concepts after passing through all of the data many times, rather than clustering based on ``first impressions'' of the data.

\paragraph{WNC} In keeping with prior work~\citep{brunet2018understanding}, use a word embedding model with context window size of ten and an embedding dimension of 100. We sample ten negative tokens per word-context pair, and we train the model for sixty epochs. Despite having more data, we do not train this model for as long, as we do not want concepts to form dense clusters or share meanings. We want the word embeddings to maintain their own individual meanings and values, as they would in a real-world setting.

\section{Matched vs Cross-Loss Influence Correlations}

\begin{figure*}[t]
\centering
\begin{subfigure}[b]{\textwidth}
\centering
        \includegraphics[width=0.4\textwidth]{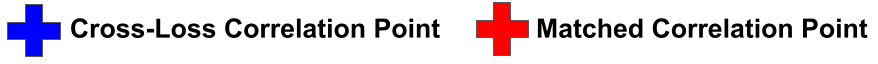}
\end{subfigure}
    \begin{subfigure}[b]{0.48\textwidth}
        \centering
        \includegraphics[width=\textwidth]{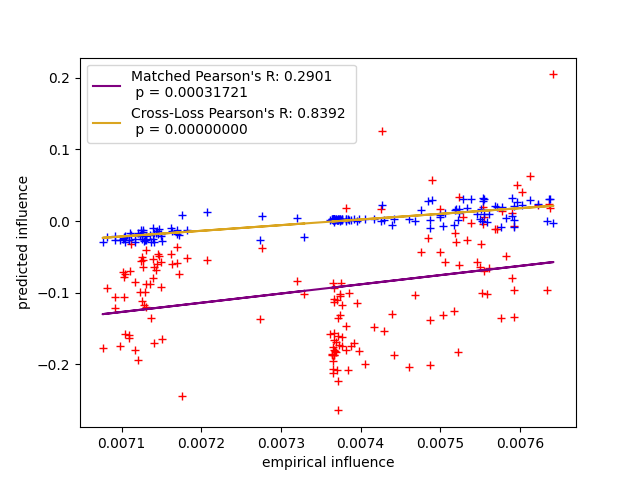}
        \label{fig:correlation_fig_a}
    \end{subfigure}
    ~~
    \begin{subfigure}[b]{0.48\textwidth}
        \centering
        \includegraphics[width=\textwidth]{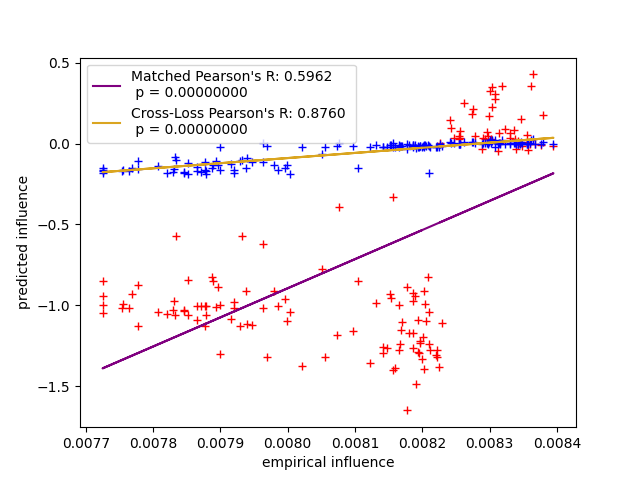}
        \label{fig:correlation_fig_b}
    \end{subfigure}
\caption{Correlation plots between cross-loss influence (blue points, orange line) and matched influence (red points, purple line) with empirical, ground-truth influence. We observe better average correlation with cross-loss influence functions.}
\label{fig:correlation_figs}
\end{figure*}

In this section, we present correlation plots between matched and cross-loss influence functions for our synthetic mixture-of-gaussians experiment (Figure \ref{fig:correlation_figs}. In each plot, empirical influence (i.e., the actual influence obtained via leave-one-out re-training) is plotted on the x-axis, and the predicted influence (according to matched or cross-loss influence functions) is depicted on the y-axis. We also plot a line of best-fit for each set of points, and present the correlation co-efficient and p-value for the Pearson's correlation in the legend.

These correlation plots show us the comparable or even improved correlation of cross-loss influence functions to empirical influence relative to matched-loss influence functions. We observe that matched-loss influence functions often provide estimates of influence that may be orders of magnitude off of ground-truth, while cross-loss influence functions are less prone to such over-estimations.